\documentclass[runningheads,a4paper]{llncs}

\usepackage{amssymb}
\usepackage{graphicx}
\usepackage[ruled,vlined]{algorithm2e}
\usepackage{amsmath}
\usepackage{changepage}
\usepackage{subfig}
\usepackage{multirow}

\usepackage{url}
\newcommand{\keywords}[1]{\par\addvspace\baselineskip
\noindent\keywordname\enspace\ignorespaces#1}
\newcommand*{\affaddr}[1]{#1} 
\newcommand*{\affmark}[1][*]{\textsuperscript{#1}}

\usepackage{fancyhdr}
\fancypagestyle{firststyle}{
    \fancyhf{} 
    
	\fancyhead[L]{\footnotesize {This work has been submitted to the IEEE for possible publication. Copyright may be transferred without notice, after which this version may no longer be accessible.}}
    
}

\begin{document}

\mainmatter  

\title{Scale Space Approximation in \\Convolutional Neural Networks for \\Retinal Vessel Segmentation}

\titlerunning{Scale Space Approximation in CNN}

%
%
\author{Kyoung Jin Noh\affmark[1] \and Sang Jun Park\affmark[1] 
\and Soochahn Lee\affmark[2\dag] \thanks{\footnotesize {This work was supported by the National Research Foundation of Korea(NRF) grant funded by the Korea government(MSIT) (NRF-2015R1C1A1A01054697).}}
}
\authorrunning{K.J. Noh, S.J. Park, S. Lee}

\institute{\affaddr{\affmark[1]Dept. Ophthalmology, Seoul Nat'l University Bundang Hospital},\\
\affaddr{\affmark[2]Dept. Electronic Eng., Soonchunhyang Univ.},\\
\email{\affmark[\dag]sclsch@sch.ac.kr}
}

%
%

\toctitle{Scale Space Approximation in CNN}
\tocauthor{Noh, Park, Lee}
\maketitle

\thispagestyle{firststyle}

\begin{abstract}
Retinal images have the highest resolution and clarity among medical images. Thus, vessel analysis in retinal images may facilitate early diagnosis and treatment of many chronic diseases. In this paper, we propose a novel multi-scale residual convolutional neural network structure  based on a \emph{scale-space approximation (SSA)} block of layers, comprising subsampling and subsequent upsampling, for multi-scale representation. Through analysis in the frequency domain, we show that this block structure is a close approximation of Gaussian filtering, the operation to achieve scale variations in scale-space theory. Experimental evaluations demonstrate that the proposed network outperforms current state-of-the-art methods. Ablative analysis shows that the SSA is indeed an important factor in performance improvement.
\end{abstract}

\section{Introduction}\label{sec:introduction}

Retinal images are widely used to observe the eye in detail, including the retina, retinal blood vessels, the optic disc, and the vitreous body. Ophthalmologists rely heavily on these images to diagnose and treat various retinal diseases including retinal tear, retinal detachment, hemorrhaging, macular degeneration, and diabetic retinopathy. It is noninvasive and simple, no radiation or pharmaceuticals are needed, and the cost for both the equipment itself and its use is low.

Retinal images are the only type of medical image that provides a clear, high resolution visualization of blood vessels in the body. Compared to other image modalities such as X-ray or CT angiography, the difference in image clarity is considerable. Thus retinal vessel analysis for early diagnosis and treatment of many chronic diseases including cardiovascular and neurovascular diseases and diabetes.


Early methods mostly applied simple thresholding to hand-designed feature descriptors, as in \cite{Frangi}. As methodologies progressed, methods with more sophisticated label inference schemes, often requiring supervised learning, together with more complex hand-designed local feature descriptors~\cite{Sofka} became more common. Vessel pixel classification methods based on supervised learning such as k-nearest neighbors~\cite{Staal} support vector machines (SVM)~\cite{Ricci} have been proposed. However, the intricacy for hand-designed feature descriptors are inevitably limited due to the limited capacity of conscience human perception. 


Deep learning methodologies can be summarized as a deep neural network (NN) structure~\cite{AlexNet,VGGNet} with weights learned by back-propagation and optimization of a loss function using stochastic gradient descent (SGD)~\cite{deeplearningbook} upon a large dataset often with ground truth supervision. While the final output of the NN is the desired outcome, and has shown excellent results as in \cite{AlexNet,VGGNet}, it has been observed that the intermediate outputs of the hidden layers of convolutional neural network (CNN) structures provide useful features. Recent works have shown that it can be used as an effective local image feature descriptor for semantic segmentation, both for general objects~\cite{FCN,UNet} and retinal vessels~\cite{DRIU,VGAN}.

\begin{figure*}[t]
\centering
\includegraphics[width=1.0\textwidth]{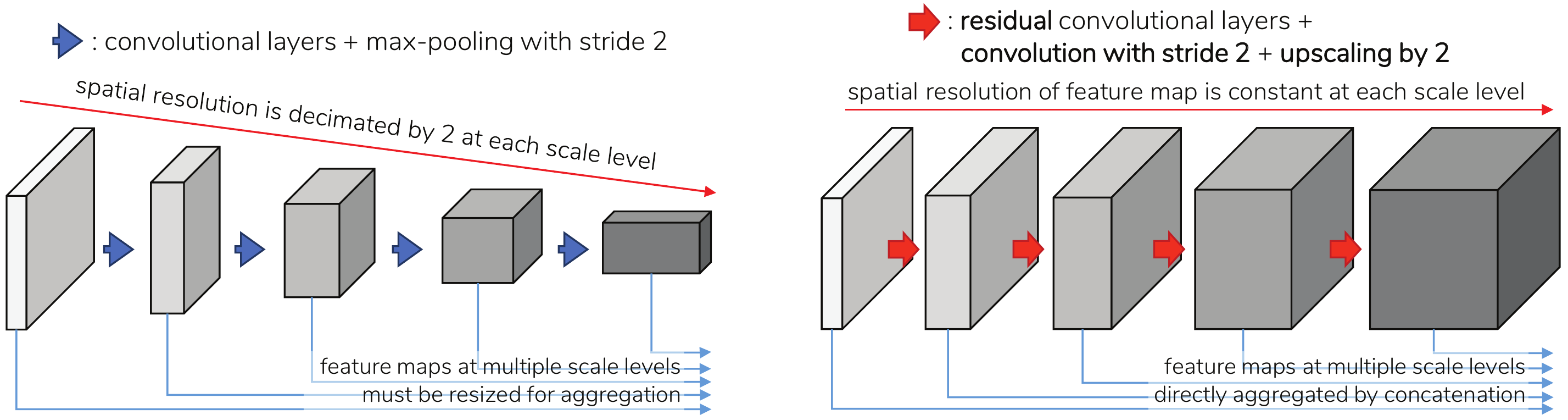}\\
	\begin{minipage}[b]{.495\linewidth}
			\centering	
	{(a) Previous}
	\end{minipage}
	\begin{minipage}[b]{.495\linewidth}
			\centering
	{(b) Proposed}
	\end{minipage}
\caption{Comparison of previous CNN structures for multi-scale feature generation, including those of \cite{UNet,DRIU}, and the proposed method. The core characteristic of the proposed retinal multi-scale residual CNN structure is the insertion of an upscaling layer following each decimating layer. This version of downscaling better adheres to the scale-space theory of \cite{Lindeberg} by better approximating the results of Gaussian blurring, and helps to give more accurate results.}
\label{fig:RMSR-CNN_struct_comp}
\end{figure*}

Deep learning methods that have been developed for retinal image segmentation~\cite{DRIU,VGAN} are commonly based on CNNs that combine multi-scale features from intermediate hidden layers, with similar preprocessing and inference schemes. Given that increasing the amount of data with ground truth is extremely difficult for retinal images, since it requires painstaking effort by an expert ophthalmologist, we believe that further research is warranted about whether there is room for improvement in the CNN structure. 

In this paper we present an analysis of the optimal CNN structure for vessel segmentation in retinal images. First, and most importantly, we analyze the multi-scale structure, comprising downsampling by subsampling or pooling and upsampling within recent CNN structures for retinal vessel segmentation. Following our analysis we present a novel and simple multi-scale structure comprising a block layer termed scale-space approximation (SSA), which is summarized in Fig.~\ref{fig:RMSR-CNN_struct_comp}. Second, we provide comparative evaluation between different multi-scale CNN structures including U-Net~\cite{UNet} and Deep Retinal Image Understanding (DRIU)~\cite{DRIU} which is based on VGGNet~\cite{VGGNet}. We find that incorporating residual convolutional blocks used in ResNet (Residual Network)~\cite{ResNet} into our proposed multi-scale CNN structure further improves accuracy. Third, we provide comparative evaluation between different numbers of layers. We also provide an ablation study regarding the combined components of the CNN configuration.  

\begin{figure*}[p]
	\begin{minipage}[b]{.4\linewidth}
			\centering	
	\includegraphics[width=1\linewidth]{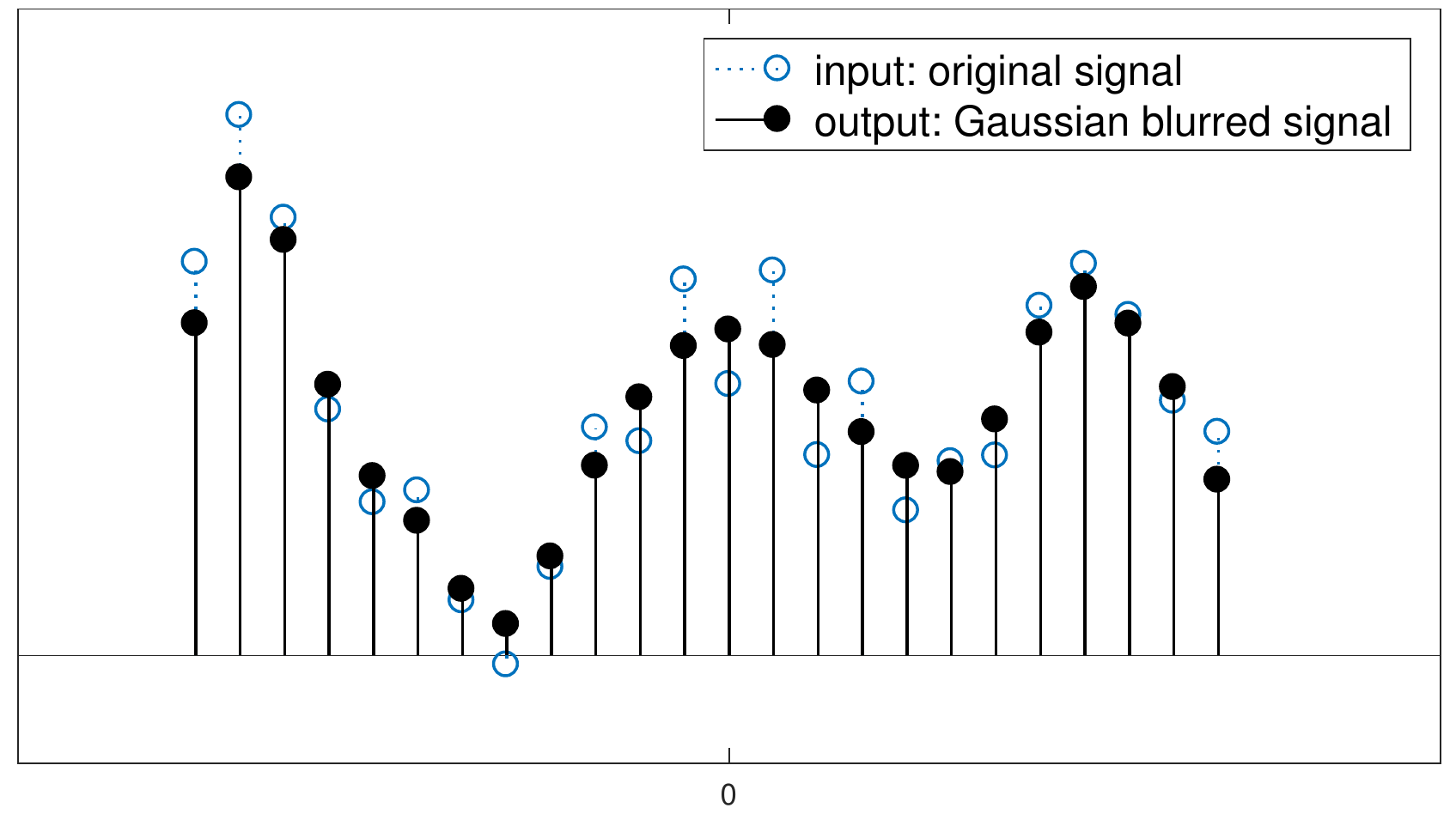}
	\end{minipage}
	\begin{minipage}[b]{.6\linewidth}
			\centering	
	\includegraphics[width=1\linewidth]{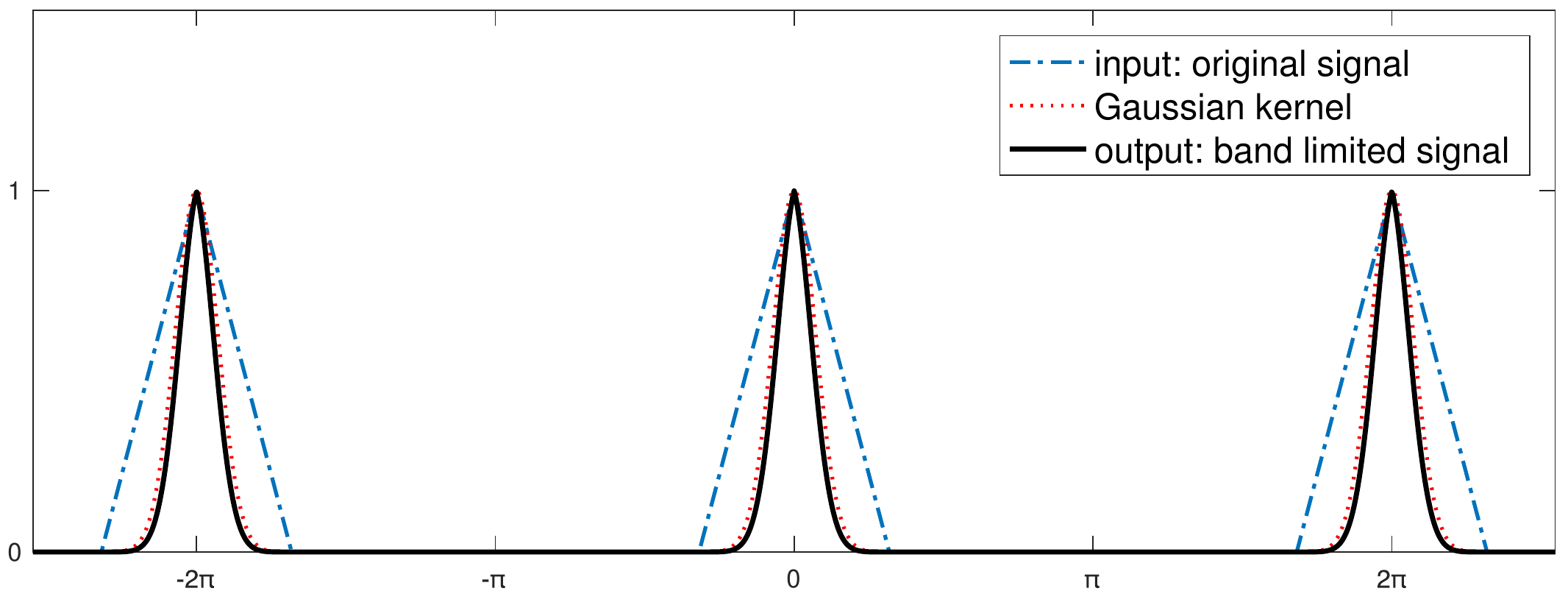}\\
	\end{minipage}
	\begin{minipage}[b]{1.0\linewidth}
	{(a) Original 1-dimensional spatial signal and its Gaussian blurred output (left) and their (simplified) frequency spectrums (right). We assume discrete Gaussian blurring with $\sigma = {{1} \over {\sqrt{2}}}$ is applied to downscale the original signal.}
	\end{minipage} \\
	\begin{minipage}[b]{.4\linewidth}
			\centering	
	\includegraphics[width=1\linewidth]{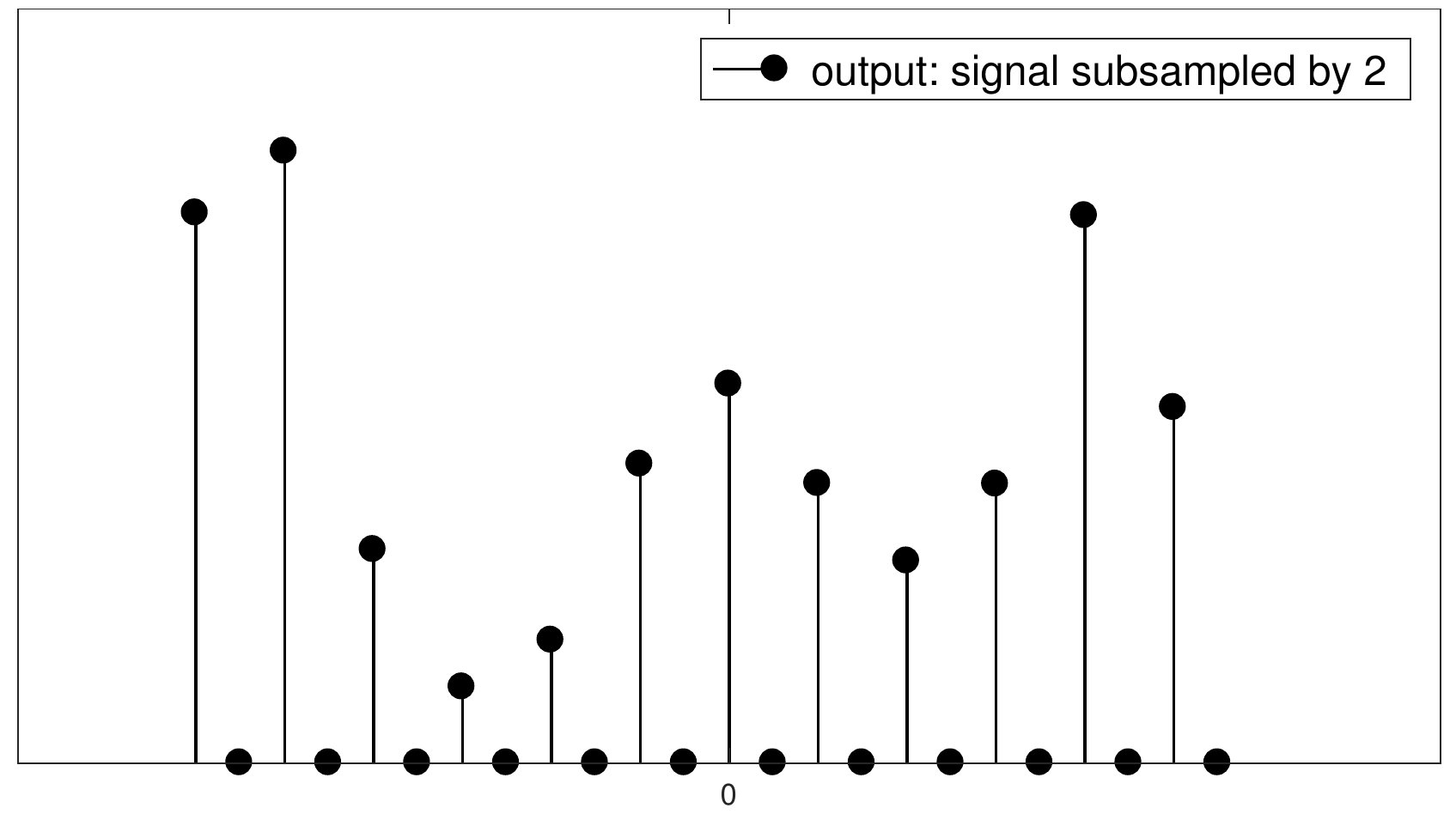}\\
	\end{minipage}
	\begin{minipage}[b]{.6\linewidth}
			\centering	
	\includegraphics[width=1\linewidth]{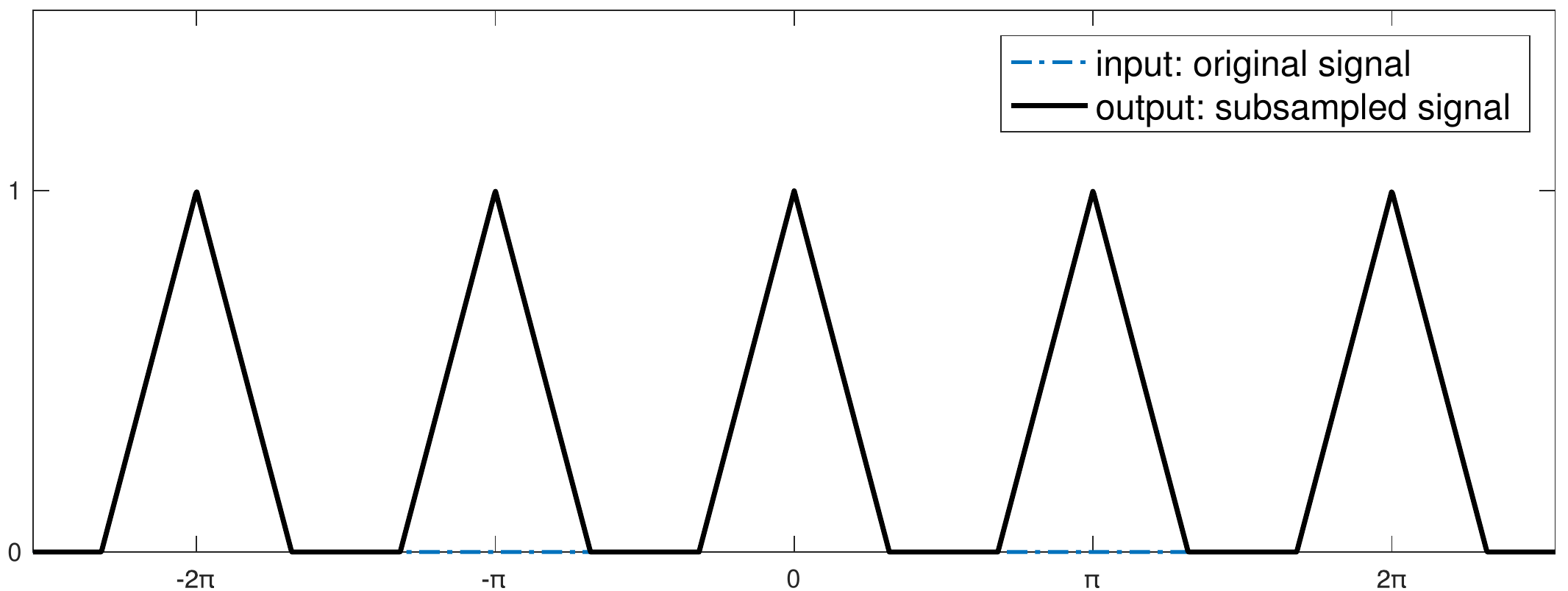}\\
	\end{minipage}
	\begin{minipage}[b]{1.0\linewidth}
		\centering{(b) 1-dimensional spatial signal (left) and its frequency spectrum (right) after subsampling by 2.}
	\end{minipage}\\
	\begin{minipage}[b]{.4\linewidth}
			\centering	
	\includegraphics[width=1\linewidth]{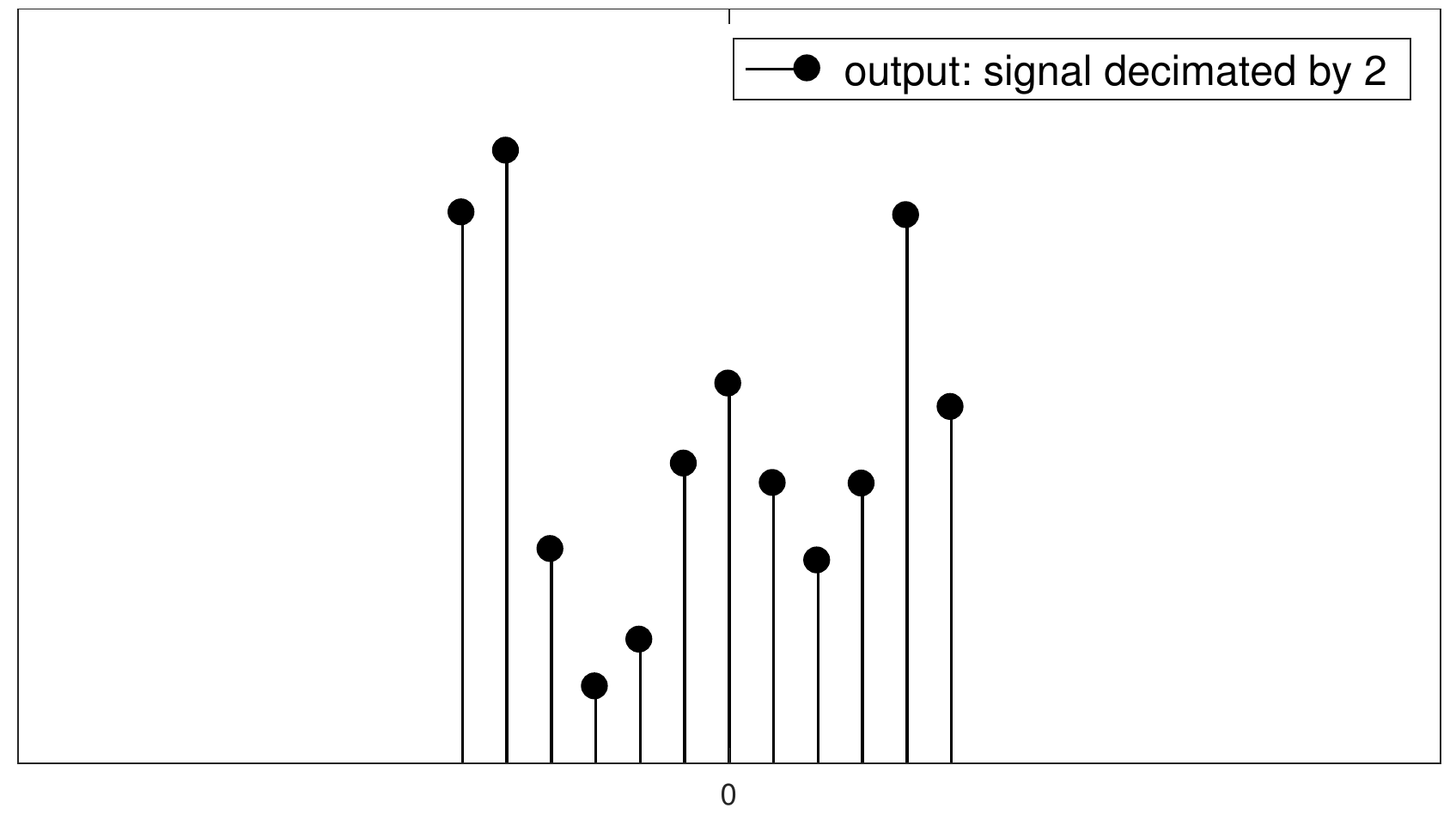}\\
	\end{minipage}
	\begin{minipage}[b]{.6\linewidth}
			\centering	
	\includegraphics[width=1\linewidth]{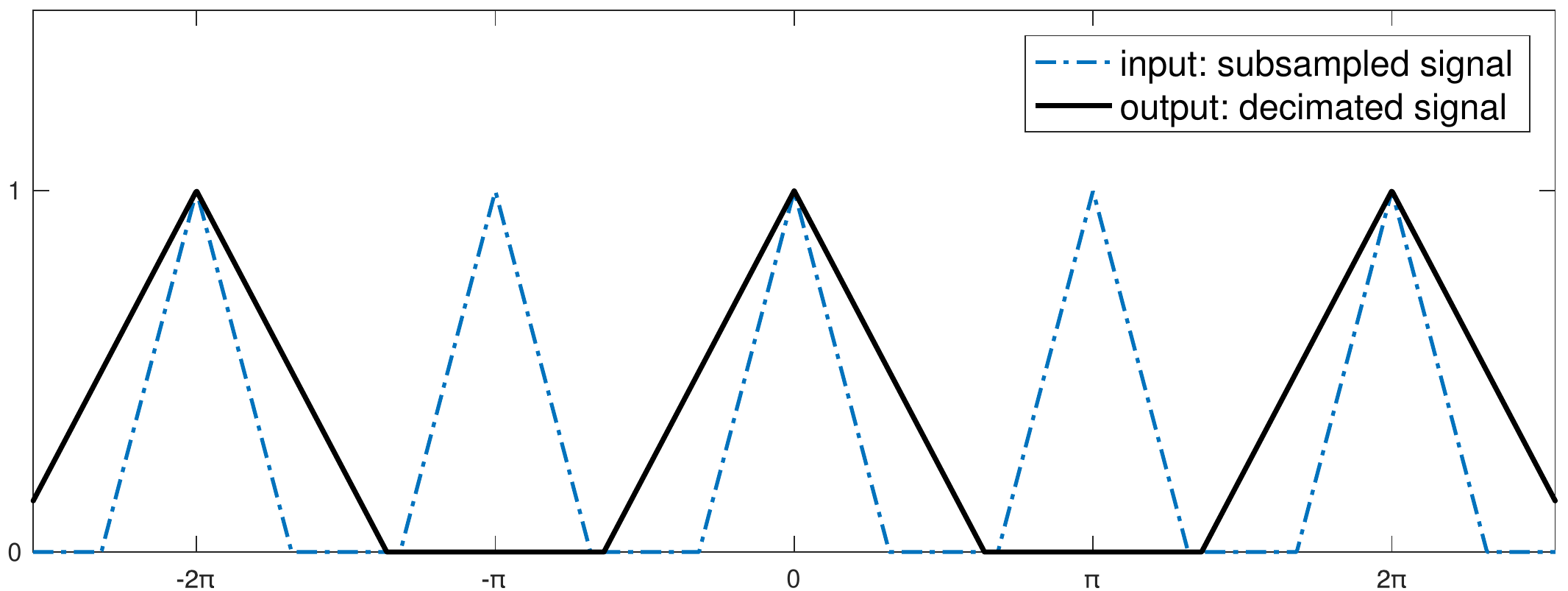}\\
	\end{minipage}
	\begin{minipage}[b]{1.0\linewidth}
		\centering{(c) The (left) spatial signal (right) frequency spectrum of the signal of (b) after decimation by 2.}
	\end{minipage}\\
	\begin{minipage}[b]{.4\linewidth}
			\centering	
	\includegraphics[width=1\linewidth]{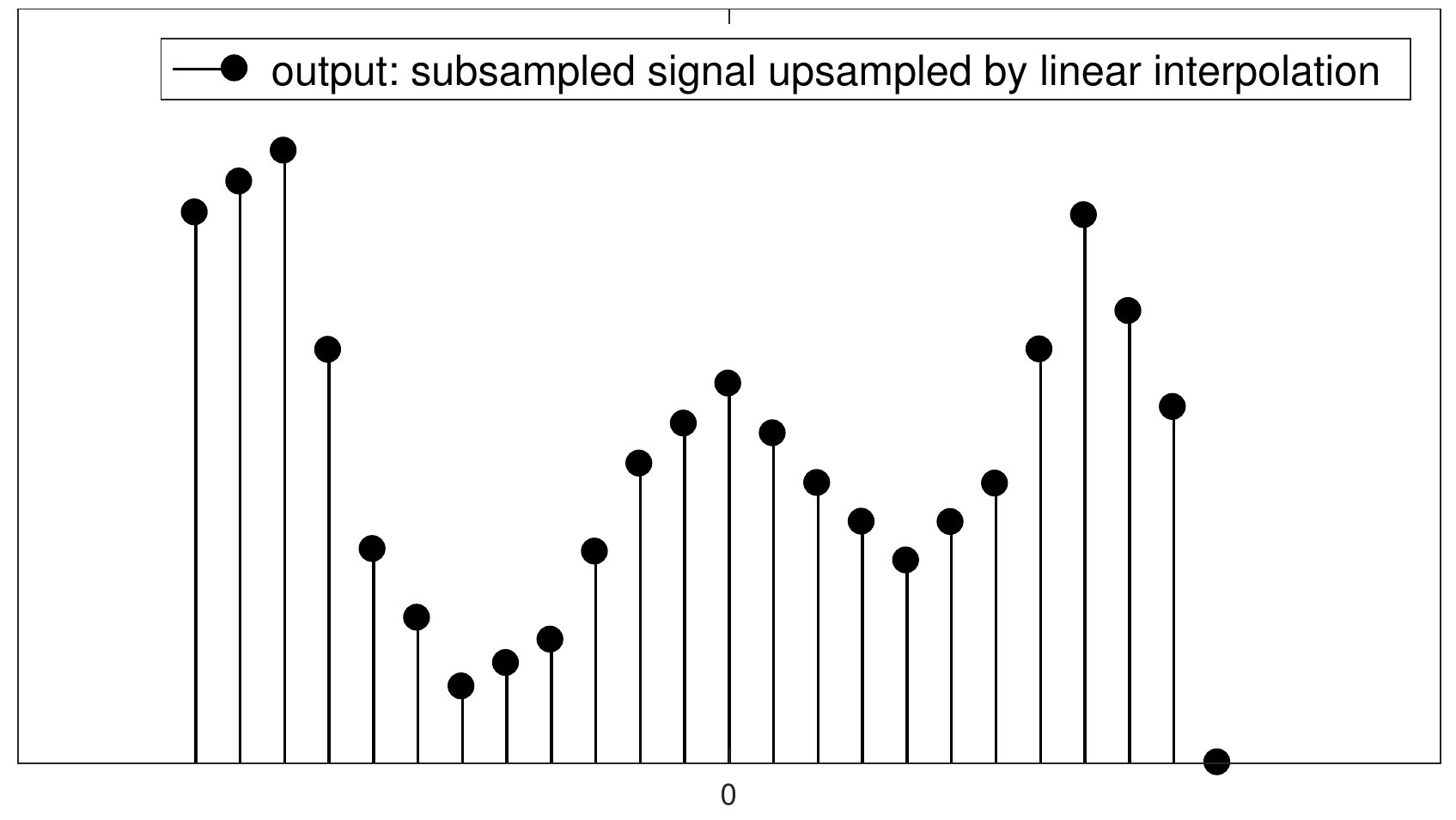}\\
	\end{minipage}
	\begin{minipage}[b]{.6\linewidth}
			\centering	
	\includegraphics[width=1\linewidth]{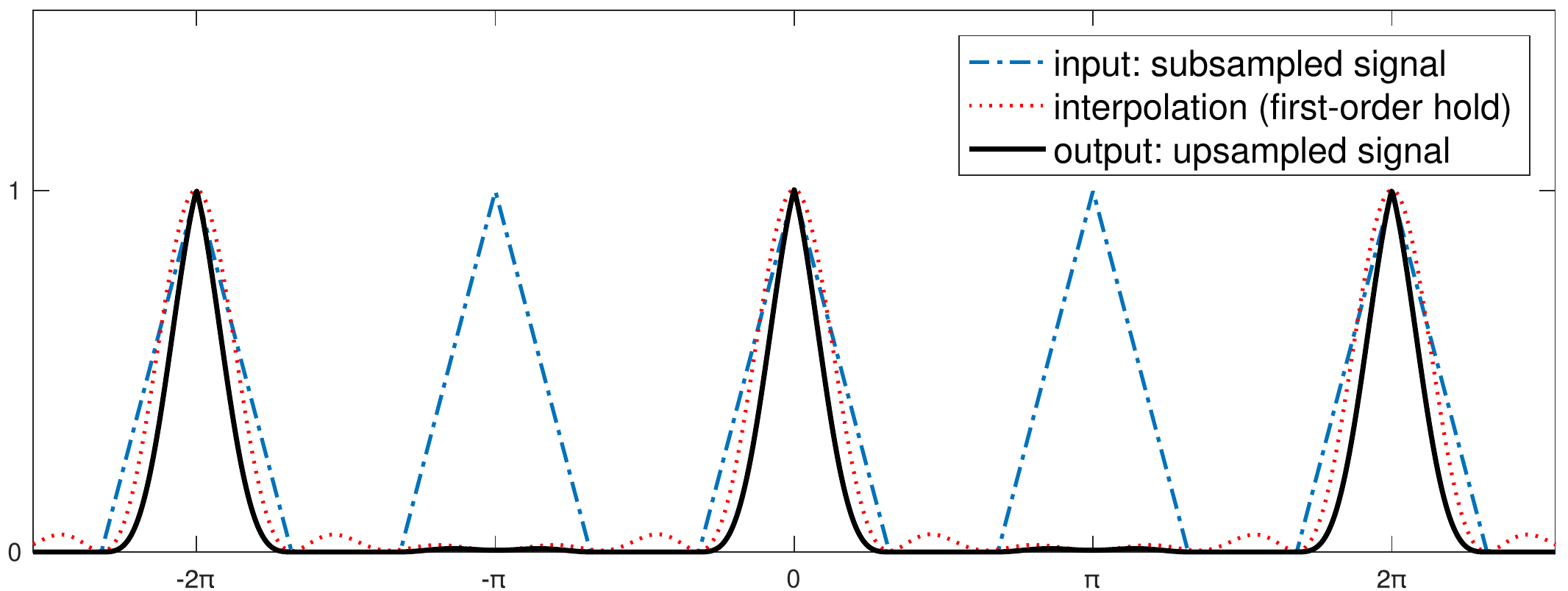}\\
	\end{minipage}
	\begin{minipage}[b]{1.0\linewidth}
		\centering{(d) The (left) spatial signal (right) frequency spectrum of the signal of (b) after linear interplation by 2.}
	\end{minipage}\\
	\caption{Frequency domain analysis of subsampling, decimation, and upsampling by bilinear interpolation for an example 1-dimensional signal. This example illustrates the difference between (c) previous downscaling methods comprising subsampling and decimation resulting, and (d) the proposed downscaling method comprising subsampling and bilinear interpolation. In all rows, the signal in the spatial domain is shown in left, and its corresponding frequency domain spectrum is shown in right. In all plots, the input (if shown) is in dashed blue, filter frequency response is in dotted red, and the output is in solid black. Figure is best viewed in color. }
	\label{fig:multi-scale_freq_analysis}
\end{figure*}

\section{Proposed Method}

\subsection{Multi-Scale Representation in CNNs with Scale Space Approximation}\label{subsec:multi_scale}

Multi-scale feature representation and inference is an especially important aspect for retinal vessel segmentation due to complex shape, high ratio of boundary pixels, and various vessel radii. In previous methods~\cite{UNet,DRIU}, multi-scale features are generated from iterative one-half downsampling implemented by max pooling per $2 \times 2$ grid with stride 2. These structures are most likely rooted in early CNN structures for classification, where pre-trained networks on the ImageNet dataset~\cite{ImageNet} are applied as initial network parameters. Here, we aim to actively determine the optimal CNN structure for representing local appearance at multiple scales. To this end, we apply the well established theory of scale space~\cite{Lindeberg} as the criterion for different structures.  

We use an a 1-dimensional signal as an example to analyze the downsampling, convolution, and upsampling in the frequency domain to gain more insight into how the original signal is altered within the CNN as shown in Fig~\ref{fig:multi-scale_freq_analysis}. The output of Gaussian blurring corresponding to downscaling in scale space theory~\cite{Lindeberg} is shown in Fig~\ref{fig:multi-scale_freq_analysis}(a). In the downscaling stage of previous methods \cite{UNet,DRIU}, the input is subsampled and decimated by 2. While subsampling and decimation jointly occur through strided pooling, their effects are different. Subsampling reduces the period of the discrete spectrum by half, as shown in Fig~\ref{fig:multi-scale_freq_analysis}(b). This is equivalent as adding an additional spectrum with but at high frequency. On the other hand, decimation stretches the frequency axis, as shown in Fig~\ref{fig:multi-scale_freq_analysis}(c). The combined effect is the doubling of the bandwidth of the input signal. Here, we can see that this does not actually downscale the signal. On the contrary, the increase in bandwidth might cause signal loss in future convolutional layers by spreading the signal spectrum.

Thus, we propose a downscaling structure that better approximates Gaussian blurring by performing upsampling by bilinear interpolation, also known as first-order hold, after downsampling. We term the block comprising the two layers of convolution with stride 2 and upsampling with ratio 2 as the scale-space approximation (SSA) block. The results of this process on the 1-dimensional signal is shown in Fig~\ref{fig:multi-scale_freq_analysis}(d). Since bilinear interpolation is represented as a triangular function in the spatial domain, its frequency response is a $sinc^{2}$ function, which is similar to the frequency response of the Gaussian, which is itself a Gaussian, both shown in red dotted lines in Figs~\ref{fig:multi-scale_freq_analysis}(a) and (d). We can also see that interpolation filters out the distortion from the previous downsampling. 


\subsection{Multi-scale Residual Network with Scale Space Approximation}\label{subsec:residual_cnn}

\begin{figure*}[tbh]
\centering
\includegraphics[width=1.0\textwidth,height=8.5cm]{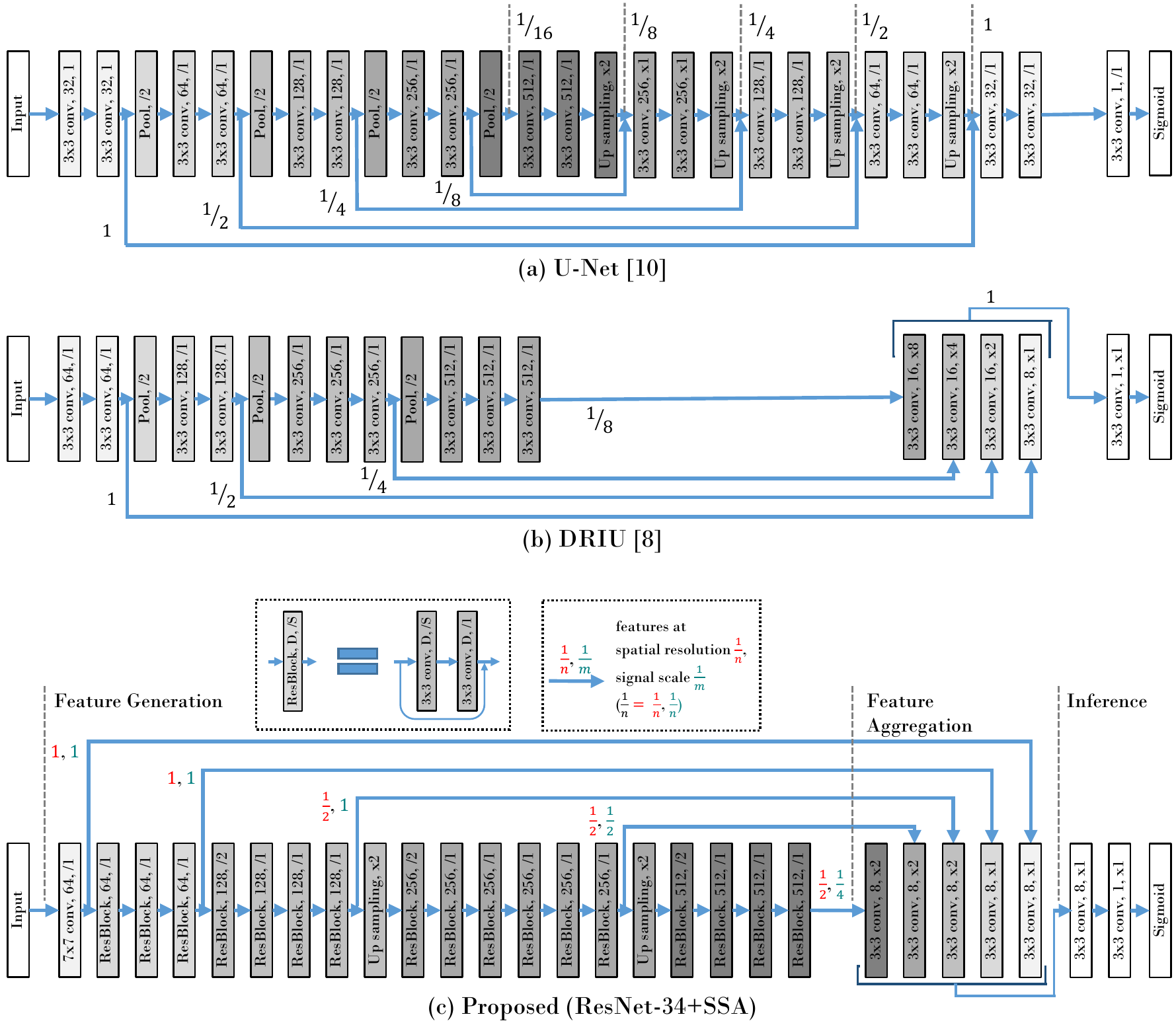}
\caption{Diagram of the complete network structures for the (a) U-Net~\cite{UNet}, (b) Deep Retinal Image Understaning (DRIU)~\cite{DRIU}, and (c) the proposed multi-scale residual network with scale space approximation (MSResNet-SSA). MSResNet-SSA has a similar structure with ResNet-34~\cite{ResNet}. Compared with ResNet-34, SSA replaces decimation by convolution with stride 2. Compared with DRIU, residual blocks replace simple convolutions. To avoid insufficient memory, we change the order of upsampling and convolutions, so that convolutions are performed in the half spatial resolution.}
\label{fig:RMSR-structure}
\end{figure*}

Based on the proposed SSA layer for multi-scale representation, we propose a novel CNN structure that shows state-of-the-art accuracy for retinal image segmentation. The proposed network combines the previous works of DRIU~\cite{DRIU} and residual networks~\cite{ResNet} as well as the SSA multi-scale representation. For a better understanding of the proposed network, we provide summarized network diagrams for U-Net~\cite{UNet}, DRIU~\cite{DRIU}, and the proposed method in Fig.~\ref{fig:RMSR-structure}.

Various new CNN structures can be constructed by replacing existing downscaling layers with the SSA for improved multi-scale representation. But two major drawbacks of the SSA, which introduces upsampling instead of decimation, must be considered. One is the reduction of receptive field, and the other is the increase in required storage. To mitigate the first issue, we adopt residual networks~\cite{ResNet} to increase the layer depth and thus the number of convolutions. For the second issue, we exchange the order of upsampling in the SSA and the convolutions, which makes the convolutions be performed in the half spatial resolution. Although this might affect the signal integrity of the convolution inputs, the effect was relatively smaller than improvement from the upsampling and residual blocks. We term the proposed network multi-scale residual network with scale space approximation (MSResNet-SSA).

\section{Results}

We experiment on the DRIVE~\cite{Staal} and STARE~\cite{STARE} datasets, comprising 40 and 20 images, respectively. Both sets contain expert annotated manual segmentations of the vessels. The segmentations of the first annotator were defined as the ground truth used to train/test the proposed MSResNet-SSA network. Segmentations from second annotator are evaluated to measure human performance. We use the train/test split as was done in \cite{DRIU}, namely, the standard division for DRIVE and the split according to which the first 10 images consist the training set and the last 10 the test set for STARE.

\begin{figure*}[tbh]
	\begin{minipage}[b]{.5\linewidth}
			\centering	
	\includegraphics[width=1\linewidth]{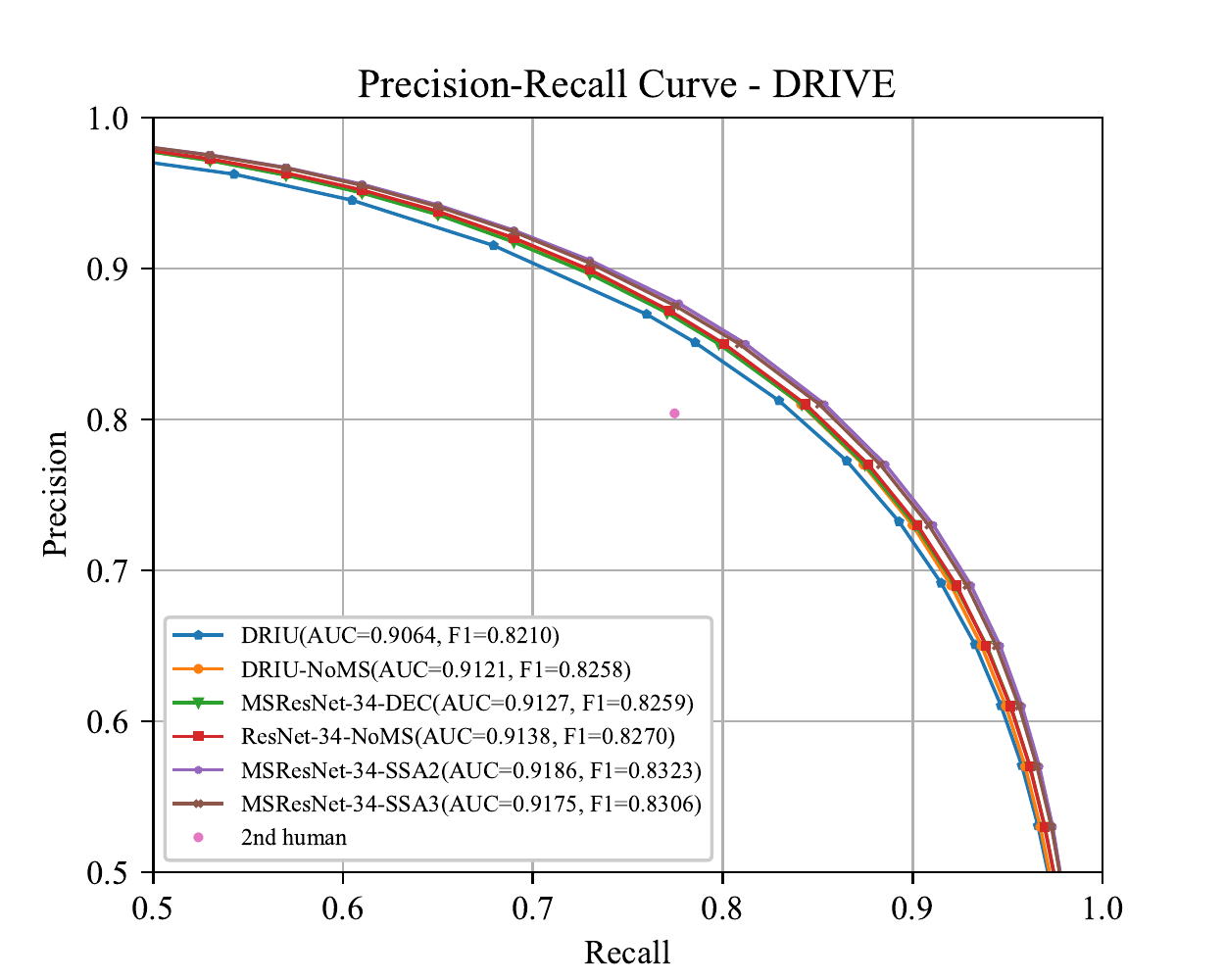}
	\end{minipage}
	\begin{minipage}[b]{.5\linewidth}
			\centering	
	\includegraphics[width=1\linewidth]{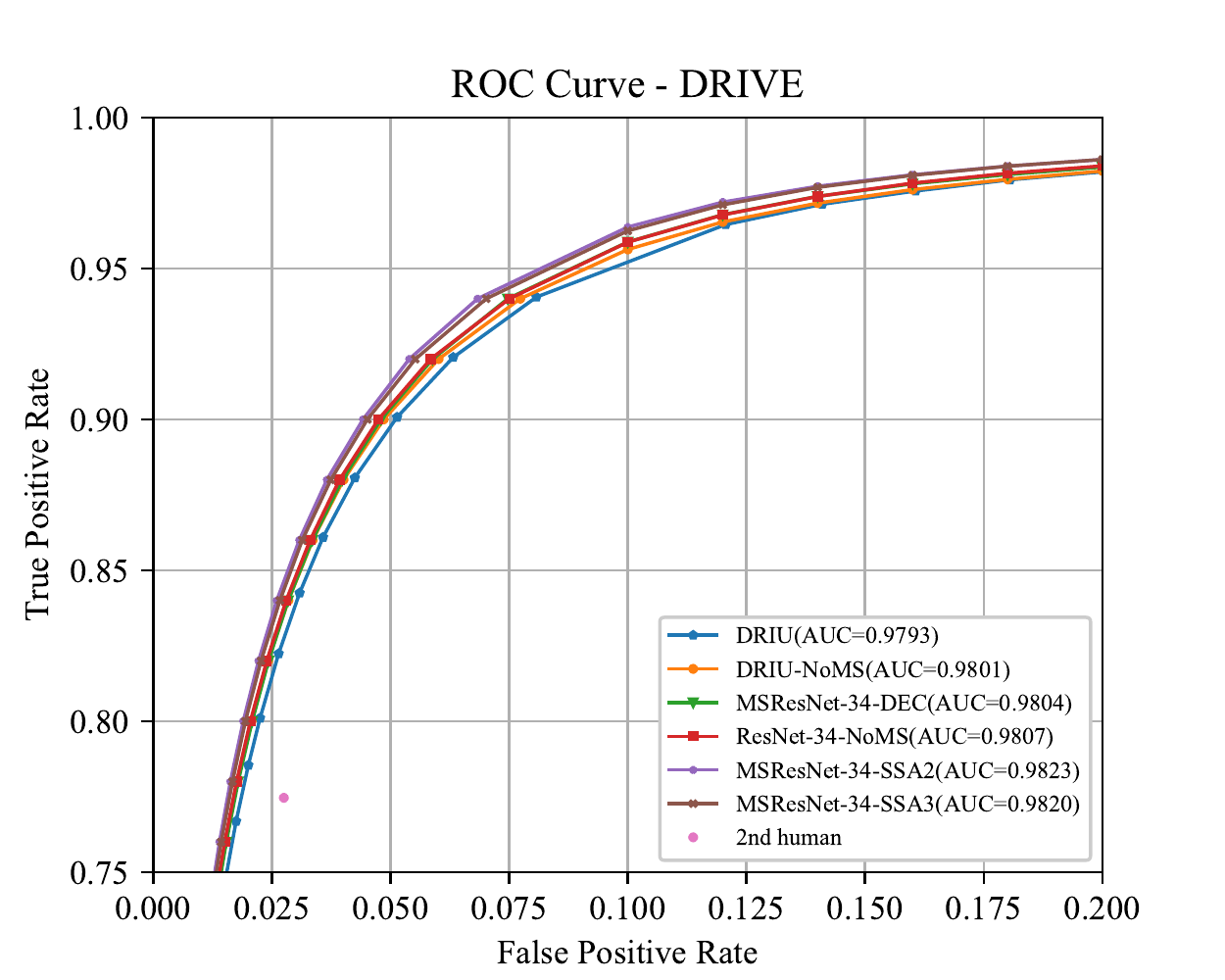}\\
	\end{minipage}
	\caption{Precision-recall (PR, left) and Receiver operation characteristic (ROC, right) curves of the proposed method MSResNet-34-SSA with various configurations, along with DRIU~\cite{DRIU} for the DRIVE~\cite{Staal} dataset. See text for details of various networks. Best viewed in color.}
	\label{fig:quant_curves_DRIVE}
\end{figure*}

\begin{figure*}[tbh]
	\begin{minipage}[b]{.5\linewidth}
			\centering	
	\includegraphics[width=1\linewidth]{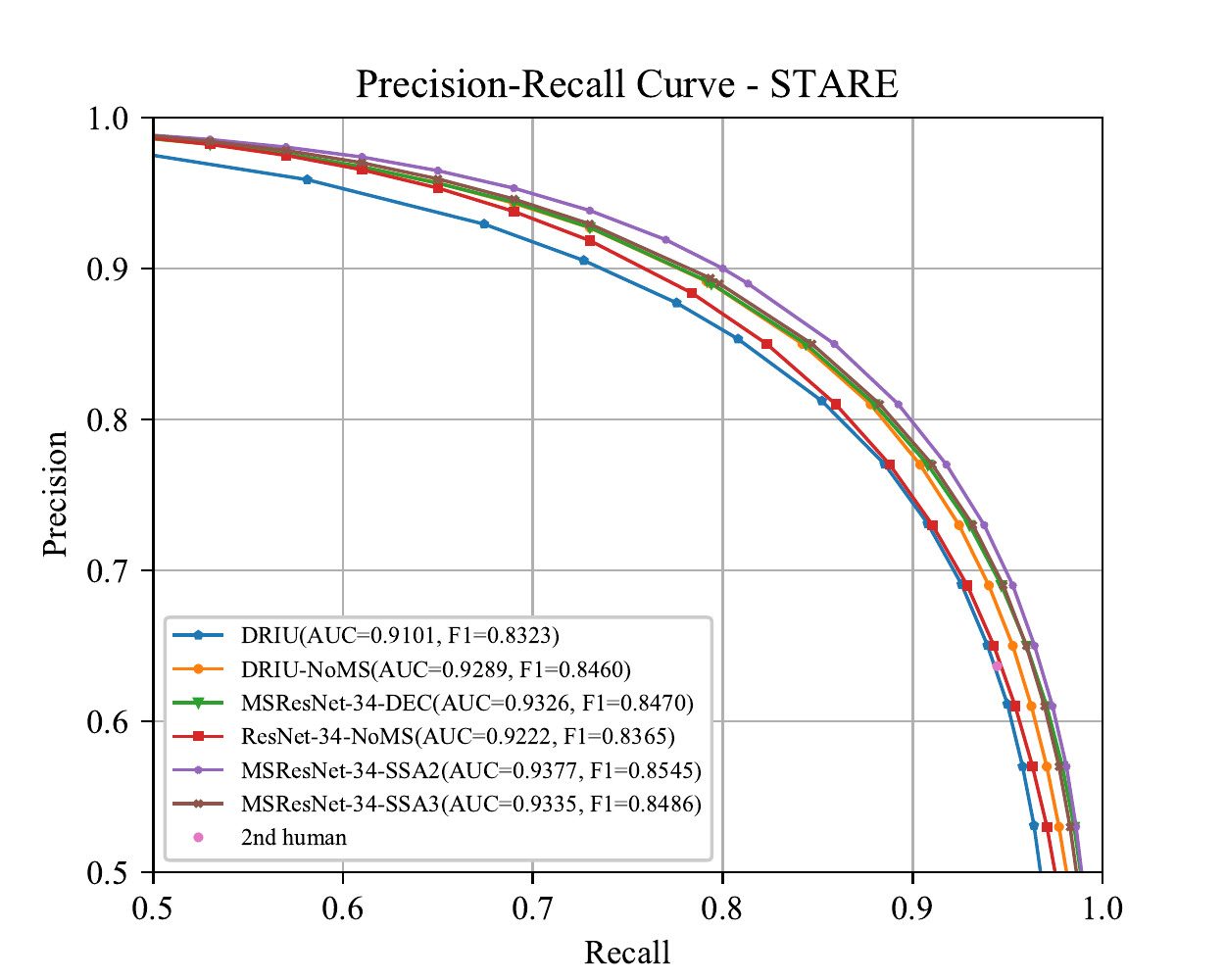}
	\end{minipage}
	\begin{minipage}[b]{.5\linewidth}
			\centering	
	\includegraphics[width=1\linewidth]{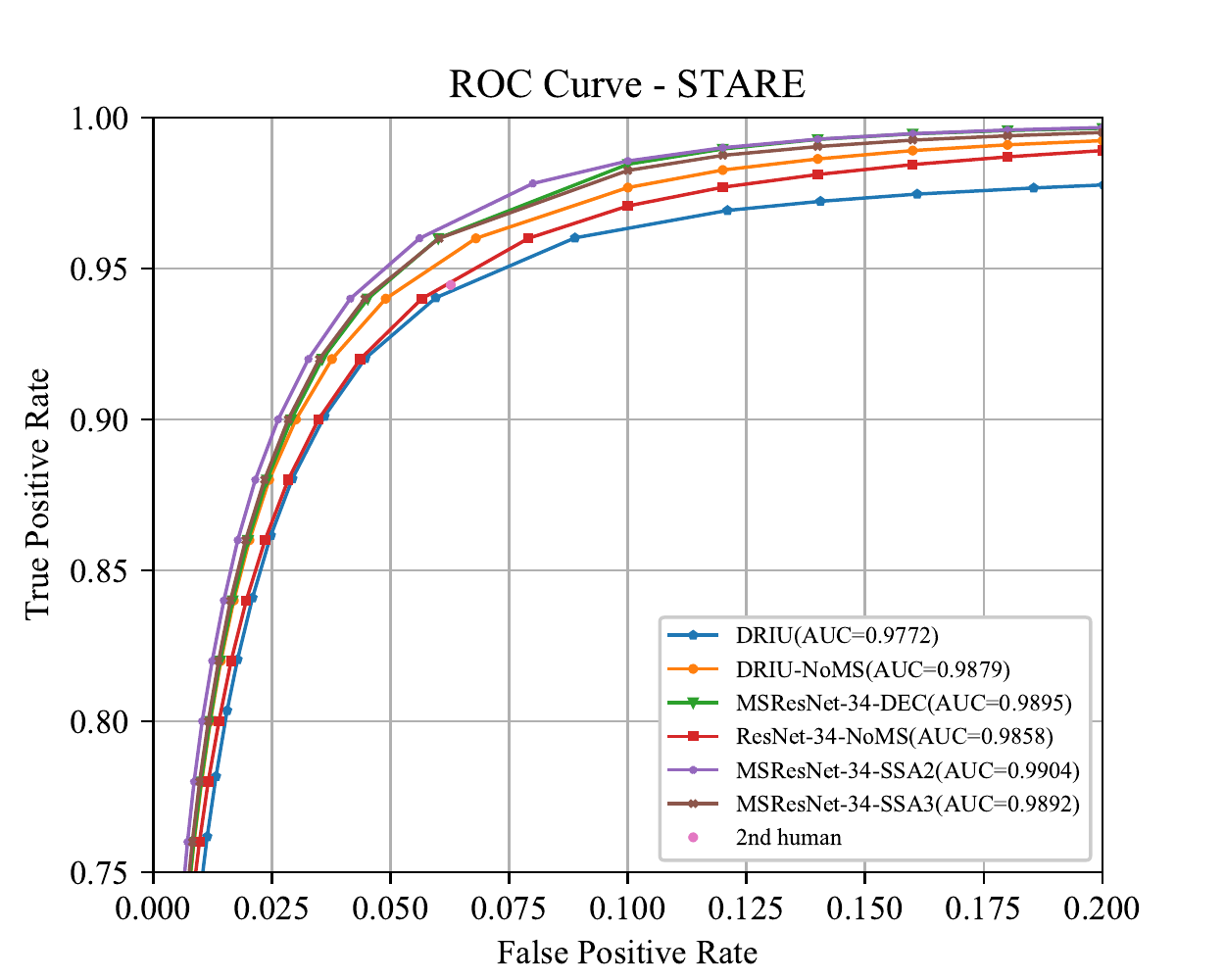}\\
	\end{minipage}
	\caption{Precision-recall (PR, left) and Receiver operation characteristic (ROC, right) curves of the proposed method MSResNet-34-SSA with various configurations, along with DRIU~\cite{DRIU} for the STARE~\cite{STARE} dataset. See text for details of various networks. Best viewed in color.}
	\label{fig:quant_curves_STARE}
\end{figure*}

We present precision-recall (PR) curves in Fig.~\ref{fig:quant_curves_DRIVE} and the receiver operating characteristic (ROC) curves in Fig.~\ref{fig:quant_curves_STARE} for the DRIVE and STARE datasets, respectively. To construct the PR and ROC curves, resulting pixel-wise probabilistic map of each image is binarized at multiple threshold values and compared to the ground truth for each method. The Dice coefficient (also known as the F1-measure and equivalent to the Jaccard index) of the optimal point in the PR curve as well as the Area-Under-Curve (AUC) values are also provided as summary measures. We provide comparisons with the current state-of-the-art method, DRIU~\cite{DRIU}, as well as comparisons with different network configurations as an ablation study. We can see that for both DRIVE and STARE datasets the proposed MSResNet-34-SSA2, which is the network with structure described in Fig.~\ref{fig:RMSR-structure}. DRIU-NoMS is the network without any subsampling or decimations, MSResNet-34-DEC is MSResNet-34 but with subsampling and decimations as in DRIU instead of the SSA, ResNet-34-NoMS is ResNet-34 without any subsampling or decimations, and  MSResNet-34-SSA3 is same as MSResNet-34-SSA2 but with an additional SSA block right after the initial convolution layer. We can see that the addition of SSA improves the results compared to previous networks with decimation, or networks with only convolutional blocks and no multi-scale consideration. One interesting result is performance drop of MSResNet-34-SSA3. Our conjecture is that aliasing occurs when subsampling early in the network since the input signal is not been sufficiently band-limited. This is avoided when subsampling after several convolutional layers because the convolutions reduced the signal bandwidth.

\begin{figure*}[tbh]
	\begin{minipage}[b]{1.0\linewidth}
			\centering	
	\includegraphics[width=1\linewidth]{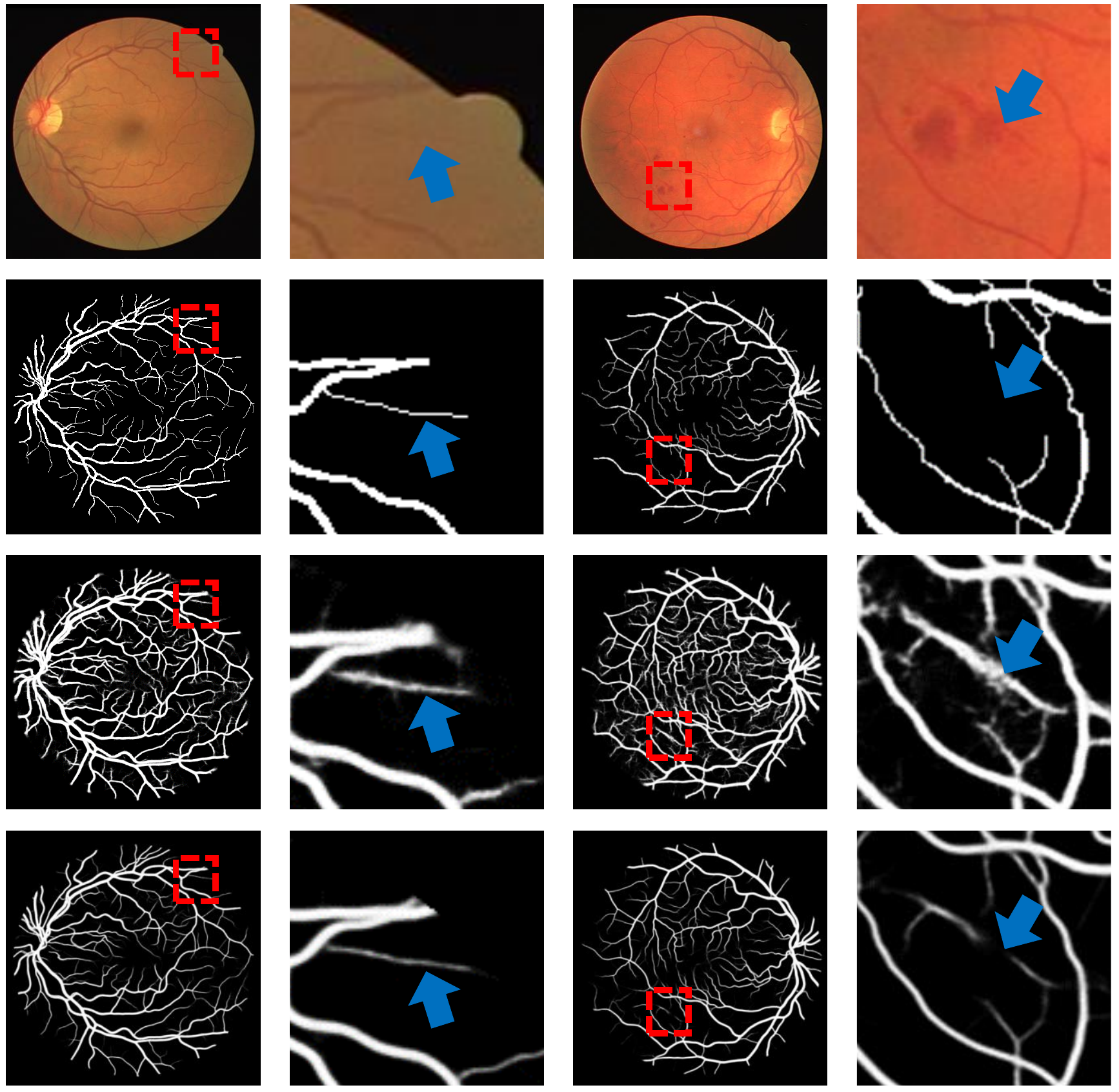}
	\end{minipage}
	\caption{Qualitative results for the DRIVE~\cite{Staal} dataset. The rows are respectively: original image (top), ground truth vessel segmentation (second), results of DRIU~\cite{DRIU} (third), and results of proposed (MSResNet-34-SSA2) method (bottom). The second and fourth columns are zoom-in portions of the corresponding first and third column images. The blue arrows point to regions where results of proposed method and the DRIU~\cite{DRIU} show major difference. We can see that the proposed method results in more accurate vessel segmentations especially for vessels that are narrow and have weak contrast.}
	\label{fig:qual_DRIVE}
\end{figure*}

\begin{figure*}[tbh]
	\begin{minipage}[b]{1.0\linewidth}
			\centering	
	\includegraphics[width=1\linewidth]{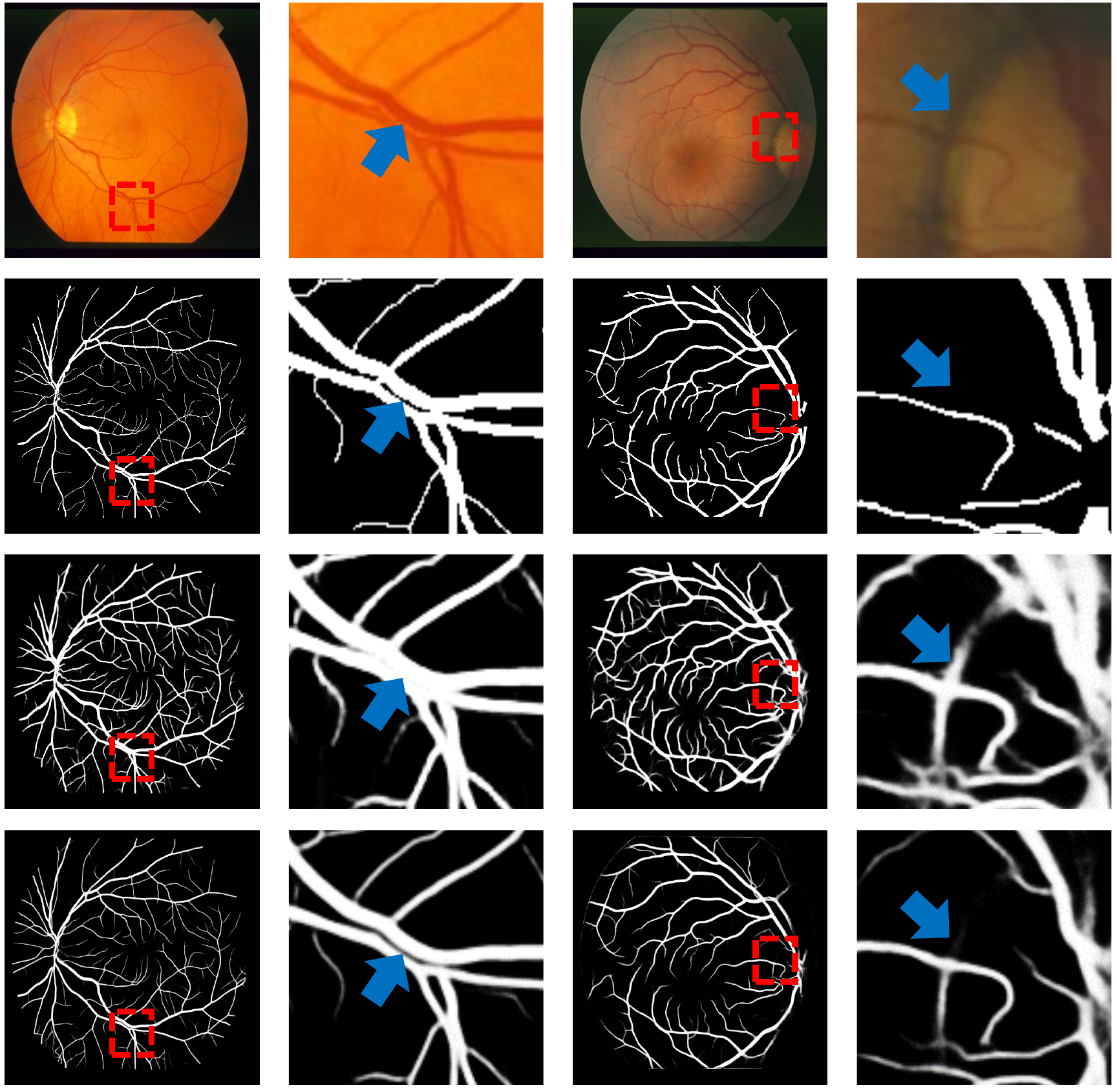}\\
	\end{minipage}
	\caption{Qualitative results for the STARE~\cite{STARE} dataset. The rows are respectively: original image (top), ground truth vessel segmentation (second), results of DRIU~\cite{DRIU} (third), and results of proposed (MSResNet-34-SSA2) method (bottom). The second and fourth columns are zoom-in portions of the corresponding first and third column images. The blue arrows point to regions where results of proposed method and the DRIU~\cite{DRIU} show major difference. We can see that the proposed method results in more accurate vessel segmentations especially for vessels that are narrow and have weak contrast.}
	\label{fig:qual_STARE}
\end{figure*}

Fig.~\ref{fig:qual_DRIVE} and Fig.~\ref{fig:qual_STARE} shows qualitative results of the proposed method along with that of the DRIU~\cite{DRIU} as a comparison. We can see that, compared to DRIU~\cite{DRIU} the proposed method results in more accurate vessel segmentations especially for vessels that are narrow and have weak contrast.

\end{document}